\documentclass{article}

\usepackage{arxiv}

\usepackage{array}
\usepackage[utf8]{inputenc} 
\usepackage[T1]{fontenc}    
\usepackage{hyperref}       
\usepackage{url}            
\usepackage{booktabs}       
\usepackage{amsfonts}       
\usepackage{nicefrac}       
\usepackage{microtype}      
\usepackage{lipsum}
\usepackage{fancyhdr}       
\usepackage{graphicx}       
\graphicspath{{media/}}     
\usepackage{amsmath}
\usepackage{algorithm}
\usepackage{algpseudocode}
\usepackage{array}

\usepackage{multirow}

\title{Beyond Full Poisoning: Effective Availability Attacks with Partial Perturbation
}

\author{
  Yu Zhe \\
  RIKEN AIP \\
  China \\
  \texttt{zhe.yu@riken.jp} \\
   \And
  Jun Sakuma \\
  RIKEN AIP; Tokyo Institute of Technology \\
  Japan\\
  \texttt{sakuma@c.titech.ac.jp} \\}

\begin{document}
\maketitle
\begin{abstract}
The widespread use of publicly available datasets for training machine learning models raises significant concerns about data misuse. Availability attacks have emerged as a means for data owners to safeguard their data by designing imperceptible perturbations that degrade model performance when incorporated into training datasets. However, existing availability attacks are ineffective when only a portion of the data can be perturbed. To address this challenge, we propose a novel availability attack approach termed Parameter Matching Attack (PMA). PMA is the first availability attack capable of causing more than a 30\% performance drop when only a portion of data can be perturbed. PMA optimizes perturbations so that when the model is trained on a mixture of clean and perturbed data, the resulting model will approach a model designed to perform poorly. Experimental results across four datasets demonstrate that PMA outperforms existing methods, achieving significant model performance degradation when a part of the training data is perturbed. Our code is available in the supplementary materials.
\end{abstract}
\section{Introduction}
\label{sec:intro}

The increasing reliance on publicly available datasets has raised a critical issue: data owners often have little control over how their data is used once it is downloaded \cite{hill2019photos}. Researchers, developers, and organizations frequently release datasets with the expectation that they will be used for specific academic or ethical purposes. However, these datasets can be freely accessed, modified, and incorporated into machine learning pipelines without the original data owners' consent. This loss of control has raised significant concerns over data misuse, privacy violations, and unauthorized commercial exploitation.

Availability attacks have been proposed as a potential solution to this problem, enabling data owners to prevent unwanted use of their datasets. These attacks work by introducing undetectable perturbations to the data, which degrade the performance of machine learning models trained on them. For example, Error-Minimizing (EM) attack \cite{huang2021unlearnable} optimizes the perturbations to minimize the classification loss obtained from the surrogate model. EM expects that the model will stop learning as the loss approaches zero. In \cite{lsp}, the authors propose generating linearly separable perturbations (LSP) as poisoning perturbations. This attack is based on the hypothesis that these linearly separable perturbations can act as shortcuts. If the model learns such shortcuts, it will produce low test accuracy on clean test data. Overall, these methods degrade model performance by introducing perturbations that either create misleading shortcuts or disrupt gradient updates, making effective learning impossible. By leveraging these techniques, data owners can effectively control the use of their data, ensuring that it cannot be used in unintended ways.

However, unlike other poisoning attacks such as backdoor attacks \cite{souri2022sleeper}, which can achieve their adversarial goal with only a small poison ratio, \textbf{existing availability attack methods work under the strong assumption of requiring a 100\% poison ratio, meaning that every sample in the dataset must be perturbed for the attack to be effective.} Prior studies \cite{huang2021unlearnable,shortcut_squ} and this work (see Table \ref{Table:baseline}) have demonstrated that when training datasets contain as little as 20\% clean data, the attack's effectiveness decrease significantly. This limitation makes current availability attacks impractical in many real-world scenarios because data exploiters can easily circumvent them by integrating even a small amount of clean data from alternative sources.

\begin{figure}[t]
    \centering
    \includegraphics[width=1.0 \columnwidth]{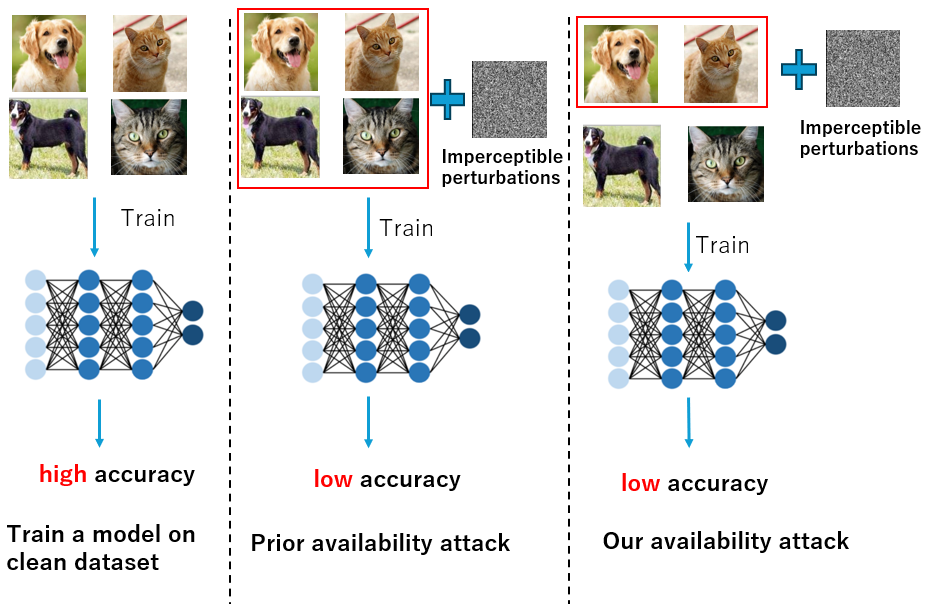} 
    \caption{Training on clean data yields high accuracy. Previous availability attacks add perturbations to all samples, making the dataset unlearnable and resulting in a model with low accuracy. Our approach requires perturbations on only part of the data to achieve unlearnability.}
    \label{fig:setup} 
\end{figure}

In this work, we challenge this fundamental assumption. We ask: \textbf{Can an availability attack still be effective when only a portion of the dataset is perturbed?} A comparison of our availability attack setting and the previous attack is shown in Figure \ref{fig:setup}. Specifically, we first analyze why existing approaches fail when the poison ratio is below 100\%. We identify that the conventional bi-level optimization framework used in prior work introduces an inherent conflict between the inner and outer optimization objectives. This conflict arises because prior methods do not explicitly consider how clean data affects the poisoning process, making it difficult to degrade model performance when clean data is present. To address this issue, we introduce \textbf{Parameter Matching Attack (PMA)}, a novel availability attack designed to remain effective even when only part of the dataset is poisoned. Specifically, our approach introduces a \textit{destination model}, which is intentionally designed to exhibit poor performance on clean test data. We then optimize the poisoned data such that the model trained on a mix of poisoned and clean samples closely resembles the poorly performing destination model. This strategy directly enforces poor model performance with considering the effect of clean data, overcoming the limitations of previous methods that fail in low-poison settings. 

Our key contributions can be summarized as follows:

\begin{enumerate}    
    \item We introduce a novel availability attack, Parameter Matching Attack (PMA). PMA is effective even when the model is trained on a mixture of perturbed and clean data. To the best of our knowledge, PMA is the first availability attack capable of causing more than a 30\% performance drop under this setting.

    \item We introduce a novel bi-level optimization framework, where the outer-level objective is uniquely designed to optimize perturbations by minimizing the distance between the resulting model and a poor-performing destination model. This distinguishes our approach from existing methods, which typically focus on direct loss manipulation. The inner-level objective follows the conventional design.

    \item Through extensive experiments conducted on four datasets compared with seven methods, we empirically show that our method is the only method that can degrade the resulting model performance by more than 30\% when the poison ratio is smaller than 100\%. 

    \item We validate our method against various countermeasures, showing that our approach maintains its effectiveness in preventing unwanted use of data even when facing adaptive countermeasures.

\end{enumerate}

\section{Background and Related Work}
\label{sec: related}

\begin{table*}[htbp]
\centering
\caption{Comparison of different poisoning attacks}
\scalebox{0.8}{\begin{tabular}{|>{\centering\arraybackslash}p{0.14\linewidth}|>{\centering\arraybackslash}p{0.14\linewidth}|>{\centering\arraybackslash}p{0.14\linewidth}|>{\centering\arraybackslash}p{0.14\linewidth}|>{\centering\arraybackslash}p{0.14\linewidth}|>{\centering\arraybackslash}p{0.14\linewidth}|}
\hline
                            & Backdoor \cite{souri2022sleeper}                            & Integrity \cite{geiping2020witches,shafahi2018poison}                    & Indiscriminate \cite{lu2023exploring,biggio2012poisoning}  & Full Availability \cite{tensorclog,deepconfuse,huang2021unlearnable,rem,tap,self_ensemble,yuan2021neural,sadasivan2023cuda} &  Partial Availability (Ours) \\ \hline
Target test sample          & Any $x$ with specific trigger $\eta$ & specific test samples $x_t$   & Any x                          & Any x                   & Any x               \\ \hline
Poison ratio                & \textless 10\%        & \textless 10\% & \textless 100\% & 100\%                   & $\leq$100\%         \\ \hline
Goal                        & $f(x+\eta) \neq y$                   & $f(x_t) \neq y$               & $f(x) \neq y$                  & $f(x) \neq y$           & $f(x) \neq y$       \\ \hline
Knowledge about test sample & $x\sim P$                            & $x_t$                          & $x\sim P$                      & $x\sim P$               & $x\sim P$           \\ \hline
Visual quality              & Usually yes                          & Usually yes                   & No                             & Yes                     & Yes                 \\ \hline
\end{tabular}}
\label{tab:summarization}
\end{table*}

Poisoning attacks involve injecting malicious data into training datasets to manipulate the behavior of resulting classification models. These attacks are typically categorized into four types: backdoor attack, integrity attack, indiscriminate data poisoning, and availability attack \cite{huang2021unlearnable, geiping2020witches, souri2022sleeper, lu2023exploring}. Backdoor attacks \cite{souri2022sleeper} involve perturbing the training data so that specific triggers in test samples lead to misclassification into a predefined category, typically affecting less than 10\% of the data (Table \ref{tab:summarization}, column 2). In an integrity attack \cite{geiping2020witches, shafahi2018poison}, the attacker modifies the training set to cause specific test samples to be misclassified while maintaining high accuracy on other samples, also usually with a low poison ratio (Table \ref{tab:summarization}, column 3). Conversely, availability attacks aim to degrade the overall test accuracy of the model by perturbing the entire training dataset. Availability attacks do not use specific triggers and aim to cause misclassification of unspecified test samples, making them more challenging than integrity and backdoor attacks \cite{huang2021unlearnable, sandoval2022autoregressive}. For this reason, the existing availability attacks require a high poison rate (usually 100\%) to maintain the effectiveness of the attack.

Another type of poisoning attack is indiscriminate data poisoning. While availability attacks and indiscriminate data poisoning both aim to reduce a model’s overall test accuracy, they differ significantly in their approach. Indiscriminate data poisoning \cite{biggio2012poisoning,lu2022indiscriminate} allows adding new samples into the dataset, which changes the total number of samples or adds unrestricted perturbations to training samples. For this reason, it takes effect even with a low poisoning ratio while the visual quality of poisoned samples is significantly low (Table \ref{tab:summarization}, column 4).  In contrast, availability attacks are allowed to add only imperceptible perturbations (e.g., limited by the $l_{p}$ norm) to the training sample to maintain visual quality without changing the total number of samples.

In this study, we focus on availability attacks. In recent years, various methods have emerged, demonstrating effectiveness when perturbations can be added to all training samples \cite{tensorclog,deepconfuse,huang2021unlearnable,rem,tap,self_ensemble,yuan2021neural}. In addition to the methods already described in Section \ref{sec:intro}, there are some studies that propose availability attacks through the use of surrogate models. Deep Confuse (DC) \cite{deepconfuse} proposes learning a generator capable of providing perturbations that maximize classification error over training trajectories of surrogate models. Targeted Adversarial Poisoning (TAP) \cite{tap} exploits a surrogate model to generate targeted adversarial examples, which can also be used for availability attacks. Self-Ensemble Protection (SEP) \cite{self_ensemble} shows that using multiple checkpoints during surrogate model training improves attack results. Neural Tangent Generalization Attacks (NTGA) \cite{yuan2021neural} use the neural tangent kernel to approximate the neural network and optimize perturbations to force misclassification. Some methods do not require surrogate models. For instance, \cite{lsp} designs linearly separable perturbations (LSP) as poisoning perturbations, and Autoregressive (AR) \cite{sandoval2022autoregressive} uses an autoregressive process to generate poisoning perturbations acting as shortcuts. Convolution-based Unlearnable Dataset (CUDA) \cite{sadasivan2023cuda} utilizes randomly generated convolutional filters to generate class-wise perturbations to achieve availability attacks.  We emphasize that existing availability attacks assume all training samples can be poisoned. Our goal is to relax this assumption. 

\section{Problem Formulation}
\subsection{Threat Model}
\label{sec: threat model}
As in previous works on availability attacks \cite{huang2021unlearnable}, our threat model defines two parties: the data owner (attacker) and the data exploiter (victim). The data owner shares image datasets on the internet. Before uploading, the data owner can add perturbations to the images to prevent unwanted use of these data. The data exploiter downloads this image dataset from the internet and uses it to train a model.

\textbf{Data owner's capability:} The data owner can add imperceptible perturbations, which are bounded by $l_p$ norm, to the data exploiter's training data. In previous availability attacks \cite{huang2021unlearnable}, the data owner is able to add perturbations to all training data. We call this setting \textit{full availability}. In this work, we relax this assumption. The data owner can only add imperceptible perturbations to part of the training data. We call this setting \textit{partial availability}. Without loss of generality, the partial availability setting can deal with the situation in which the data exploiter obtains data from other data sources and merges them with the data owner's data. The data from all sources are supposed to be i.i.d samples of an underlying distribution. Subsequently, the data exploiter trains a model from scratch on a mixture of perturbed data and clean data by empirical risk minimization. The success of the data owner is measured by the resulting model's accuracy on clean test images, with lower accuracy indicating greater success.

\textbf{Data owner's knowledge:} The data owner has limited knowledge about the data exploiter’s training process. Specifically, the data owner is unaware of the exploiter’s model architecture, initialization, and training details. However, the data owner can train a local model, which serves as a surrogate model for generating perturbations. In the partial availability setting, the data exploiter’s training data consists of both clean and perturbed samples. In practice, the data owner does not have access to the specific clean data used by the data exploiter. Instead, we assume the data owner has access to an additional dataset drawn from the same underlying data distribution as their original dataset, which serves as a proxy for the clean data used by the data exploiter. This assumption is reasonable because data owners typically have knowledge of their data distribution, allowing them to obtain additional samples consistent with their original dataset. Even if obtaining new samples is not feasible, the data owner can instead reserve a subset of their original dataset as a proxy for the clean data used by the data exploiter.

\subsection{Problem Setup}
\label{sec: setup}
Let $\mathcal{X}$ be the sample space, and $\mathcal{Y}$ be the label space. Let data $(x,y)$ follow an underlying distribution $\mathcal{P}$ over $\mathcal{X} \times \mathcal{Y}$. We consider the classification problem to predict $y \in \mathcal{Y}$ given $x\in \mathcal{X}$ where $(x,y) \sim \mathcal{P}$. 

\textit{Full availability:} in this setting, the data owner has a clean dataset $\mathcal{D}_{\text{cl}}$. The data owner is able to add perturbations $\delta=\left\{\delta_i\right\}_{i=1}^N$ to $\mathcal{D}_{\text{cl}}=\left\{\left(x_i, y_i\right)\right\}_{i=1}^{N}$. Then, the data owner will release $\mathcal{D}_{\text{poi}}= \{ (x + \delta,y) | (x,y) \in  \mathcal{D}_{\text{cl}} \}$ in public. The objective to construct $\delta$ is that: \begin{align}
&\quad \quad \quad \max _{\delta} \mathbb{E}_{(x, y) \sim \mathcal{P}} \mathcal{L}\left(F(x ; \theta^{*}( \mathcal{D}_{\text{poi}} )), y\right) \label{eq:full_o}\\
&\text{s.t.} \quad \scalebox{0.9}{$\theta^{*}( \mathcal{D}_{\text{poi}} ) =\underset{\theta}{\min } \sum_{(x,y) \in \mathcal{D}_{\text{cl}}} \mathcal{L}\left(F\left(x+\delta; \theta\right), y\right)$}. \label{eq:full_availability}
\end{align}

On the other hand, the data exploiter's objective is to find a classifier $F: \mathcal{X} \rightarrow \mathcal{Y}$ that minimizes the loss function: $\underset{\theta}{\min } \quad \mathbb{E}_{(x,y) \sim \mathcal{P}} \mathcal{L}\left(F\left(x; \theta\right), y\right)$. If the data exploiter is not aware that the publicly released data is affected by the availability attack, it trains the model with $\mathcal{D}_{\text{poi}}$ as it is: \begin{equation} 
\theta^{*}(\mathcal{D}_{\text{poi}}) = \underset{\theta}{\min } \sum_{(x,y) \in \mathcal{D}_{\text{cl}}} \mathcal{L}\left(F\left(x+\delta; \theta\right), y\right).
\label{eq:erm}
\end{equation}

\textit{Partial availability:} in this setting, the data exploiter collects both $\mathcal{D}_{\text{poi}}$ from the poisoned data source and additional clean data $\mathcal{D}_{\text{extra}}$ from a clean data source that is unknown to the data owner. Here, $\mathcal{D}_{\text{extra}} \sim \mathcal{P}$.

If the data owner is aware that the data exploiter behaves in the partial availability setting, the data owner will modify the strategy for optimizing perturbations to respond to the data exploiter's strategy adaptively. We assume that the data owner can have additional clean data $\tilde{\mathcal{D}}_{\text{extra}} \sim \mathcal{P}$ to simulate $\mathcal{D}_{\text{extra}}$ used by the data exploiter. A straightforward strategy is extending eq.\ref{eq:full_o} and eq. \ref{eq:full_availability} into the following: \begin{align}
&\quad \quad \quad \max _{\delta} \mathbb{E}_{(x, y) \sim \mathcal{P}} \mathcal{L}\left(F(x ; \theta^{*}(\tilde{\mathcal{D}}_{\text{extra}} \cup \mathcal{D}_{\text{poi}} )), y\right) \label{eq:outter_o} \\
&\text{s.t.} \quad \scalebox{0.9}{$\theta^{*}(\tilde{\mathcal{D}}_{\text{extra}} \cup \mathcal{D}_{\text{poi}} ) = \underset{\theta}{\arg\min} [ \sum\limits_{(x,y) \in \tilde{\mathcal{D}}_{\text{extra}}} \mathcal{L}\left(F\left(x; \theta\right), y\right)$} \notag \\
&\quad \quad \quad \quad \quad \quad \quad \quad \scalebox{0.9}{$+ \sum\limits_{(x,y) \in \mathcal{D}_{\text{cl}}} \mathcal{L}\left(F\left(x+\delta; \theta\right), y\right) ]$} \label{eq:inner_o}
\end{align}

Then, the data exploiter aims to obtain a classifier by: \begin{equation} 
\begin{split}
\theta^{*}(\mathcal{D}_{\text{extra}} \cup \mathcal{D}_{\text{poi}} ) = &  \underset{\theta}{\min } \sum_{(x,y) \in \mathcal{D}_{\text{extra}}} \mathcal{L}\left(F\left(x; \theta\right), y\right) \\
& + \sum_{(x,y) \in \mathcal{D}_{\text{cl}}} \mathcal{L}\left(F\left(x + \delta; \theta\right), y\right) .
\end{split}
\label{eq:erm}
\end{equation}

\section{Method}
\label{sec: method}
\subsection{Limitation in previous solutions}
\label{sec:limitation}
\cite{deepconfuse,gradient_matching,yuan2021neural} proposed attack methods for solving eq.\ref{eq:full_o} and eq. \ref{eq:full_availability}. A simple modification of these attack methods for the partial availability setting would also provide a straightforward solution for the eq. \ref{eq:outter_o} and eq. \ref{eq:inner_o}. However, our empirical investigation revealed that such extensions could not attain sufficient attack effectiveness in the partial availability setting. We show the empirical results in the supplementary.

We introduce a high-level hypothesis to explain why the straightforward solution does not work well for the partial availability setting. For the partial availability attack, when optimizing eq. \ref{eq:inner_o}, the training procedure involves minimization of $\sum_{(x,y) \in \tilde{\mathcal{D}}_{\text{extra}} }\mathcal{L}\left(F\left(x; \theta\right), y\right)$ with $\tilde{\mathcal{D}}_{\text{extra}} \sim \mathcal{P}$. On the other hand, when optimizing eq. \ref{eq:outter_o}, it involves maximization of generalization loss with clean data $\mathbb{E}_{(x, y) \sim \mathcal{P}} \mathcal{L}\left(F\left(x ; \theta^*\left(\tilde{\mathcal{D}}_{\text {extra }} \cup \mathcal{D}_{\text {poi }}\right)\right), y\right)$. Since data are randomly taken from $\mathcal{P}$ in both, the minimization of eq. \ref{eq:inner_o} and the maximization of eq. \ref{eq:outter_o} partially conflict. Such a conflict could make solving the problem inherently difficult.

\subsection{Our Strategy}
\label{sec: our_setup}

\begin{figure*}[h]
    \centering
    \includegraphics[width=0.9 \textwidth]{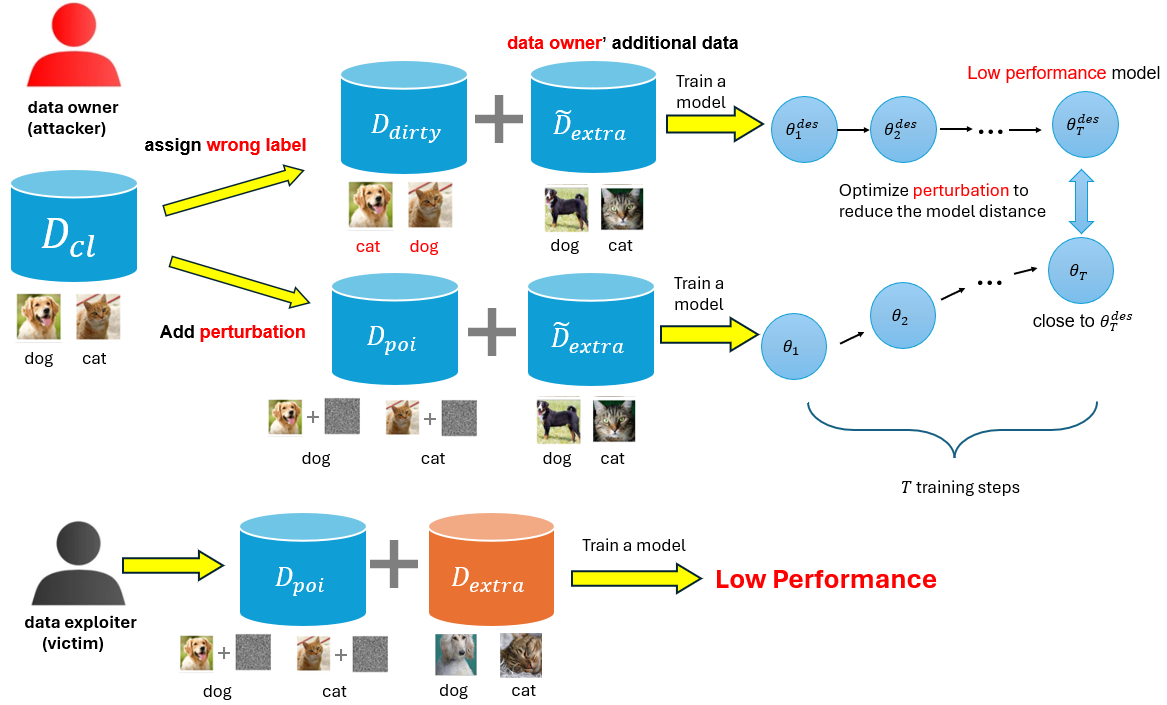} 
    \caption{The data owner assigns wrong labels to $D_\text{cl}$ resulting in $D_\text{dirty}$; adds perturbations to $D_{\textbf{cl}}$ resulting in $D_{\text{poi}}$. The perturbations are optimized so that the model trained on the mixture of $D_{\text{poi}}$ and $\tilde{\mathcal{D}}_{\text{extra}}$ gradually approaches the parameters of the model trained on $D_{\text{dirty}}$ and $\tilde{\mathcal{D}}_{\text{extra}}$. The data owner controls datasets marked in blue; The dataset marked in orange is unknown to the data owner. } 
    \label{fig:illu} 
\end{figure*}

In this section, to overcome the above difficulty, we introduce a novel and generalized formulation for the poisoning attack, framed as a bi-level optimization problem. Specifically, we introduce a destination model, which is a model designed intentionally to perform poorly with clean data. With this, we aim to optimize perturbations so that the resulting model is close to the destination model. A high-level illustration is shown in Figure \ref{fig:illu}.

Formally, we convert eq. \ref{eq:outter_o} and eq. \ref{eq:inner_o} into the following:
\begin{align}
& \min _{\delta} d\left(\theta^{*}(\tilde{\mathcal{D}}_{\text{extra}}\cup\mathcal{D}_{\text{poi}}), \theta^{*}_{\text{des}}\right) \label{outter_pm} \\
& \text{s.t.} \quad \theta^{*}(\tilde{\mathcal{D}}_{\text{extra}} \cup \mathcal{D}_{\text{poi}} ) = \underset{\theta}{\arg\min} \left[ \sum\limits_{(x,y) \in \tilde{\mathcal{D}}_{\text{extra}}} \mathcal{L}\left(F\left(x; \theta\right), y\right) \right. \notag \\
& \quad \quad \quad \quad \quad \quad \quad \quad \left. + \sum\limits_{(x,y) \in \mathcal{D}_{\text{cl}}} \mathcal{L}\left(F\left(x+\delta; \theta\right), y\right) \right] \label{eq:inner_pm}
\end{align}
where $d$ is a distance measurement between two model parameters. $\theta^{*}_{\text{des}}$ is the model parameter of the destination model. With the minimization of this distance, we expect the model trained on a mixture of clean and poisoned data to perform closely to the destination model with low performance. In this bi-level optimization problem, the two minimization terms do not have directly conflicting components, so we expect that this objective can realize the partial availability attack with sufficient attack effectiveness.

\textbf{Design of $\theta^{*}_{\text{des}}$:} By setting $\theta^{*}_{\text{des}}$ as the model parameters of a model with low performance, the goal of the availability attack is expected to be attained. We consider a heuristic design of the destination model that achieves good attack performance:

{\small
\begin{align}
\theta^{*}_{\text{des}} &= \theta^{*}(\tilde{\mathcal{D}}_{\text{extra}} \cup \mathcal{D}_{\text{dirty}} ) \notag  = \underset{\theta}{\arg\min} \left[ \sum\limits_{(x,y) \in \tilde{\mathcal{D}}_{\text{extra}}} \mathcal{L}\left(F\left(x; \theta\right), y\right) \right. \notag \\
&\quad + \left. \sum\limits_{(x,y) \in \mathcal{D}_{\text{cl}}} \mathcal{L}\left(F\left(x; \theta\right), g(y) \right) \right]
\label{eq:erm_des}
\end{align}
} where $g(y)$ is a function to generate a dirty label, we change the originally correct labels to uniformly random incorrect labels. As shown in \cite{zhang2024badlabel}, this $g(y)$ is a good way to cause model performance degradation. By using this $g(y)$, we construct a dirty label dataset $\mathcal{D}_{\text{dirty}}=\{(x,g(y)) | (x,y) \in  \mathcal{D}_{\text{cl}} \}$. There also exist other strategies to construct a destination model with low performance, such as a model with random weights. We show experimental results with other destination models in the supplementary.

\subsection{Sub-goal: full knowledge setting}
\label{sec: subgoal}
Recall that the data owner's final goal is to attain attack without any knowledge about $\mathcal{D}_{\text{extra}}$. As a sub-goal, we first achieve our goal in a relatively easy setting by assuming that the data owner can know the extra clean data used by the data exploiter, that is, $\tilde{\mathcal{D}}_{\text{extra}} = \mathcal{D}_{\text{extra}}$, we call this scenario the full knowledge setting. In this case, the data owner can better simulate the behavior of the data exploiter. Then, the objective of optimizing perturbations is achieved by replacing $\tilde{\mathcal{D}}_{\text{extra}}$ with $\mathcal{D}_{\text{extra}}$ in eq. \ref{outter_pm}, \ref{eq:inner_pm} and \ref{eq:erm_des}.

Directly solving this optimization problem is difficult since each step of the perturbation update requires the computation of $\theta^{*}(\mathcal{D}_{\text{extra}}\cup\mathcal{D}_{\text{poi}})$. Getting this parameter requires a whole training process, which is prohibitively expensive. Inspired by \cite{trajetory}, we, therefore, consider an approximate solution, the detailed algorithm is shown in Algorithm \ref{alg:main}. During the model training, we alternatively optimize perturbation to reduce the distance between the two intermediate model parameters trained on $\mathcal{D}_{\text{extra}}\cup\mathcal{D}_{\text{dirty}}$ and on $\mathcal{D}_{\text{extra}}\cup\mathcal{D}_{\text{poi}}$ in each step, as well as training the two model by $\mathcal{D}_{\text{extra}}\cup\mathcal{D}_{\text{poi}}$ or $\mathcal{D}_{\text{extra}}\cup\mathcal{D}_{\text{dirty}}$ (line 7-13 in algorithm \ref{alg:main}). Also, as mentioned in the introduction, the perturbations need to be imperceptible. Hence we bound the size of perturbations by the $l_p$ norm. We add an additional constraint that $\|\delta\|_{\infty} \leq \epsilon$ into the above formulation. This constraint is satisfied by the projected gradient minimization (line 9-13 in algorithm \ref{alg:main}). Also, since the data owner cannot know $F$, we use a surrogate model $F'$ to replace $F$ in the algorithm. 

 \begin{algorithm}[t]
\caption{Availability attack via parameter matching}\label{alg:cap}
\begin{algorithmic}[1]
\Require: known clean dataset $\mathcal{D}_{\text{extra}}$, dataset that can be perturbed $\mathcal{D}_{\text{cl}}$, number of total iteration $T$, number of steps for optimizing perturbation $J$, step length for optimization perturbation $\beta$, the bound for perturbation $\epsilon$,learning rate $\alpha$, a surrogate model $F'$, a function to generate dirty label $g()$.
\State{Initialize the perturbation $\delta = \left\{\delta_i\right\}_{i=1}^N$ that will be added into $\mathcal{D}_{\text{cl}}=\left\{(x_i,y_i)\right\}_{i=1}^{N}$, each $\delta_{i}$ correspond to $x_i$ in $\mathcal{D}_{\text{cl}}$}
\State{Initialize $\theta_0$ and $\theta^{\text{des}}_{0}$}
 \For{t=0,...,T-1}
    \State{Sample $m$ sample-label pairs from $\mathcal{D}_{\text{extra}}$, $(x^{\text{extra}}, y^{\text{extra}})$}
    \State{Sample $n$ sample-label pairs from $\mathcal{D}_{\text{cl}}$, $(x^{\text{cl}}, y^{\text{cl}})$}
    \State{Select $n$ perturbation $\delta^{\text{cl}}$ that corresponds to $x^{\text{cl}}$} 
    \State{$\theta_{t+1} \leftarrow \theta_{t}-\alpha \nabla_{\theta_t}( \mathcal{L}\left(F'( x^{\text{cl}} + \delta^{\text{cl}};\theta_t) , y^{cl}\right)+\mathcal{L}\left(F'( x^{\text{extra}};\theta_t) , y^{\text{extra}}\right))$}
    \State{$\theta^{\text{des}}_{t+1} \leftarrow \theta^{\text{des}}_{t}-\alpha \nabla_{\theta_t}( \mathcal{L}\left(F'( x^{\text{cl}};\theta^{\text{des}}_{t}) , g(y^{\text{cl}})\right)+\mathcal{L}\left(F'( x^{\text{extra}};\theta^{\text{des}}_{t} ) , y^{\text{extra}}\right))$}
    \For{$j=0,...,J-1$} 
        \State{$G=\nabla_{x^{\text{cl}}+\delta^{\text{cl}}} d (\theta^{\text{des}}_{t}, \theta^{\text{des}}_{t+1}, \theta_{t+1})$}
        \State{$\delta^{\text{cl}}=\Pi_\epsilon\left(x^{\text{cl}} + \delta^{\text{cl}} -\beta \cdot \operatorname{sign} (G) \right)-x^{\text{cl}}$} 
        \State{$\delta^{\text{cl}}=\text{Clip}( x^{\text{cl}}+\delta^{\text{cl}},0,1 ) - x^{\text{cl}}$}             
    \EndFor
\EndFor
\State{return $\mathcal{D}_{\text{poi}}=\left\{\left(x_i + \delta_i, y_i\right)\right\}_{i=1}^N$}    
\end{algorithmic}
\label{alg:main}
\end{algorithm}

We follow \cite{trajetory} and use the normalized squared distance to measure the distance between the model parameters. $$
d \left(\theta^{\text{des}}_{t}, \theta^{\text{des}}_{t+1}, \theta_{t+1}\right)=\frac{||\theta^{\text{des}}_{t+1}-\theta_{t+1}||^2}{||\theta^{\text{des}}_{t+1}-\theta^{\text{des}}_t||^2}    
$$
As training proceeds, the model parameters change less and less, so using by $||\theta^{\text{des}}_{t+1}-\theta^{\text{des}}_t||^2$ for normalization, this loss can be encouraged to be effective even after a certain phase of training has been carried out.

\subsection{Final Goal: sampling oracle model}
\label{sec: unknown}
Finally, we consider the more realistic scenario, $\tilde{\mathcal{D}}_{\text{extra}} \neq \mathcal{D}_{\text{extra}}$, we call this scenario sampling oracle setting. In this situation, the data owner collects alternative data $\tilde{\mathcal{D}}_{\text{extra}}$ in the original data distribution in place of $\mathcal{D}_{\text{extra}}$. The objective corresponding to this setting has been described by eq. \ref{outter_pm}, eq. \ref{eq:inner_pm}, eq. \ref{eq:erm_des}. Then, we can use Algorithm \ref{alg:main} immediately for this setting by replacing $\mathcal{D}_{\text{extra}}$ with $\tilde{\mathcal{D}}_{\text{extra}}$.

\section{Experiment}
\subsection{Setup}
\textbf{Datasets and models.} We utilized four datasets to evaluate our proposal: SVHN \cite{svhn}, CIFAR-10 \cite{c10}, CIFAR-100 \cite{c10}, and a 100-class subset of ImageNet \footnote{We use 20\% of the first 100 class subset, follows \cite{huang2021unlearnable}}\cite{imagenet}. If not mentioned specifically, we used a three-layer ConvNet to serve as the surrogate model for generating perturbations on the data owner's side. On the data exploiter's side, we used ConvNet as the target model on SVHN, and used ResNet-18 \cite{resnet} as the target model on CIFAR-10, CIFAR-100, and ImageNet. We trained all models for 100 epochs. We used Adam with a training rate of 0.01.

\textbf{Simulate the partial availability setting:} To simulate the full knowledge setting, we divided the dataset into two parts: a subset to be perturbed, which corresponds to $\mathcal{D}_{\text{cl}}$, and another subset of remaining clean data, which corresponds to $\mathcal{D}_{\text{extra}}$. For the sampling oracle setting, the dataset was divided into three subsets: $\mathcal{D}_{\text{cl}}$ (clean data to be perturbed), $\mathcal{D}_{\text{extra}}$ (extra clean data for the data exploiter), $\tilde{\mathcal{D}}_{\text{extra}}$ (extra clean data for the data owner). Since the data owner cannot know the number of samples in $\mathcal{D}_{\text{extra}}$, we set the number of samples in $\tilde{\mathcal{D}}_{\text{extra}}$ was equal to that in $\mathcal{D}_{\text{cl}}$. We then used Algorithm \ref{alg:main} to generate a poisoned dataset based on these subsets and trained the model on a mixture of poisoned and clean data for the data exploiter. The number of samples in each subset was adjusted according to the poison ratio. More details are shown in the supplementary.

\textbf{Comprasion methods:} We compare our results with major existing availability attacks \cite{huang2021unlearnable,rem,lsp,tap,deepconfuse,sandoval2022autoregressive,sadasivan2023cuda}. We used these methods to poison a portion of training data and trained the model on a mixture of the poisoned and remaining clean data. We evaluate our proposal and competitive methods by measuring the classification accuracy of trained models on clean test data. All results are averaged over ten runs.

\begin{table*}[t]
\caption{Results of the partial (poison ratio=80-20\%) availability attack in the sampling oracle setting when perturbation size ranges from (8/255, 16/255, 25/255). The attack performance was evaluated with classification accuracy (\%) on SVHN, CIFAR-10, CIFAR-100, and ImageNet datasets. Lower means better attack performance.}
\centering
\scalebox{0.85}{\begin{tabular}{c|cccc|cccc|cccc|cccc}
\hline
           & \multicolumn{4}{c|}{SVHN} & \multicolumn{4}{c|}{CIFAR-10} & \multicolumn{4}{c|}{CIFAR-100} & \multicolumn{4}{c}{ImageNet} \\ \hline
Ratio      & 20   & 40   & 60   & 80   & 20    & 40    & 60    & 80    & 20     & 40    & 60    & 80    & 20    & 40    & 60    & 80   \\ \hline
Clean only & \multicolumn{4}{c|}{96.1} & \multicolumn{4}{c|}{94.2}     & \multicolumn{4}{c|}{77.4}      & \multicolumn{4}{c}{62.1}     \\ \hline
8/255      & 94.3 & 92.8 & 92.9 & 92.7 & 93.5  & 92.3  & 91.3  & 90.9  & 76.1   & 75.3  & 74.3  & 73.6  & 61.4  & 61.5  & 60.8  & 59.9 \\ \hline
16/255     & 93.9 & 90.5 & 84.4 & 81.3 & 92.7  & 91.1  & 83.9  & 78.0  & 74.9   & 71.8  & 63.2  & 57.6  & 60.2  & 59.9  & 54.3  & 50.2 \\ \hline
25/255     & 89.2 & 82.7 & 71.0 & 65.7 & 88.4  & 72.0  & 64.1  & 60.2  & 71.5   & 61.0  & 43.8  & 36.4  & 56.8  & 50.6  & 42.3  & 37.3 \\ \hline
\end{tabular}}
\label{table:diff_norm}
\end{table*}

\begin{table*}[t]
\caption{Results of the full (poison ratio=100\%) and partial (poison ratio=80-40\%) availability attack in the full knowledge setting (Ours-Full) and sampling oracle setting (Ours-Oracle). The attack performance was evaluated with classification accuracy (\%) on four datasets. Lower means better attack performance.}
\centering

\scalebox{0.75}{\begin{tabular}{l|cccc|cccc|cccc|cccc}
\hline
  Dataset                & \multicolumn{4}{c|}{SVHN}                                    & \multicolumn{4}{c|}{CIFAR-10}                                & \multicolumn{4}{c|}{CIFAR-100}                               & \multicolumn{4}{c}{ImageNet}                                 \\ \hline
  Poison ratio                & 40\%          & 60\%          & 80\%          & 100\%        & 40\%          & 60\%          & 80\%          & 100\%        & 40\%          & 60\%          & 80\%          & 100\%        & 40\%          & 60\%          & 80\%          & 100\%        \\ \hline
Clean data only   & \multicolumn{4}{c|}{96.1}                                    & \multicolumn{4}{c|}{94.1}                                    & \multicolumn{4}{c|}{77.4}                                    & \multicolumn{4}{c}{62.1}                                     \\ \hline
EM \cite{huang2021unlearnable}               & 93.1          & 91.8          & 90.9          & \textbf{6.4} & 91.9          & 90.1          & 90.9          & 19.2         & 75.3          & 74.0          & 71.8          & 6.9          & 61.5          & 59.8          & 59.3          & 26.7         \\ \hline
REM \cite{rem}              & 92.2          & 90.9          & 87.5          & 11.9         & 93.5          & 94.1          & 93.8          & 21.8         & 76.6          & 75.4          & 73.4          & 8.9          & 62.4          & 61.7          & 61.8          & 9.2 \\ \hline
DC \cite{deepconfuse}               & 94.3          & 94.4          & 93.4          & 23.5         & 92.9          & 91.3          & 86.4          & 14.8         & -             & -             & -             & -            & -             & -             & -             & -            \\ \hline
TAP \cite{tap}              & 92.5          & 91.7          & 89.6          & 9.7          & 91.9          & 90.6          & 87.2          & \textbf{8.2} & 76.0          & 73.5          & 69.7          & 7.8          & 60.2          & 60.4          & 57.5          & 21.8         \\ \hline
LSP \cite{lsp}              & 91.7          & 90.1          & 88.2          & 6.8          & 92.8          & 91.4          & 88.1          & 19.1         & 74.5          & 73.6          & 70.2          & 3.1          & 59.8          & 58.4          & 56.6          & 26.3         \\ \hline
AR \cite{sandoval2022autoregressive}               & 93.6          & 93.0          & 90.7          & 6.8          & 93.2          & 91.8          & 88.4          & 11.8         & 73.2          & 71.9          & 68.6          & \textbf{3.0} & -             & -             & -    & -            \\ \hline
CUDA \cite{sadasivan2023cuda}               &  92.6         &  91.3         &   90.8       &      13.7     &    92.7       &   91.6        &   88.6       &  18.5      &      72.8     &   71.4        &   70.0       &   12.7 &      58.2        &    57.5          &  56.3   &   \textbf{8.9}          \\ \hline

Ours-Full            & \textbf{81.5} & \textbf{73.4} & \textbf{62.5} & 19.4         & \textbf{71.5} & \textbf{63.8} & \textbf{57.8} & 14.8         & \textbf{59.6} & \textbf{41.0} & \textbf{33.7} & 13.2         & \textbf{49.7} & \textbf{44.3} & \textbf{39.8} & 14.1         \\ \hline
Ours-Oracle            & \textbf{82.7} & \textbf{71.0} & \textbf{65.7} & 23.2         & \textbf{72.0} & \textbf{64.1} & \textbf{60.2} & 19.4         & \textbf{61.0} & \textbf{43.8} & \textbf{36.4} & 9.4          & \textbf{50.6} & \textbf{42.3} & \textbf{37.3} & 23.7         \\ \hline
\end{tabular}}
\label{Table:baseline}
\end{table*}
\subsection{Evaluate perturbation size and poison ratio}
\label{sec:diff_size}
We first evaluate the effectiveness of our method across varying perturbation sizes and poison ratios for the partial availability setting, generating perturbations with $\ell_{\infty}-$norm as 8/255, 16/255, and 25/255, and varying the poison ratio from 20\% to 80\%. When perturbations are bounded by $\ell_{\infty}-$norm with 8/255 (4th row in Table \ref{table:diff_norm}) or 16/255 (5th row in Table \ref{table:diff_norm}), our proposal achieves only limited performance degradation, this aligns with findings from \cite{shortcut_squ,sadasivan2023cuda}, which demonstrate that existing full availability attacks \cite{huang2021unlearnable,rem,lsp,tap,deepconfuse,sandoval2022autoregressive,sadasivan2023cuda} also fail to achieve partial availability with these smaller sizes, this is because partial availability attacks are challenging and require relatively larger perturbation sizes. The 6th row in Table \ref{table:diff_norm} shows that our proposal works well when the perturbation size is 25/255, so we adopt this size for our proposal and comparison methods \footnote{Only CUDA follows the hyperparameters from the \cite{sadasivan2023cuda}, since the size of the perturbation cannot be directly controlled.} in all subsequent experiments for a fair comparison. The visualization of images with different perturbation sizes is shown in the supplementary. Additionally, the first column in each dataset of Table \ref{table:diff_norm} empirically indicates that our method requires at least a 40\% poison ratio to be effective, as a 20\% ratio achieves insufficient results. Therefore, we set a minimum poison ratio of 40\% to achieve effective attack performance in our experiments. 

\textbf{Discussion:} Unlike backdoor attacks, which can achieve their objectives with a poison ratio as low as 10\%, our approach requires a 40\% poison ratio. However, this should not be considered excessive, as our goal is fundamentally different. Rather than embedding hidden behaviors into the model, our method focuses on limiting the usability of a dataset. From the perspective of a data exploiter, reducing the proportion of poisoned data below 40\% would necessitate obtaining a substantial amount of clean data from alternative sources, potentially exceeding the size of the original dataset. If such large-scale data collection were feasible, the exploiter could bypass the dataset entirely, rendering the attack scenario irrelevant. Since this level of data collection is impractical, our requirement of a 40\% poison ratio is already sufficient to restrict unwanted dataset usage while remaining feasible in real-world settings.

\subsection{Evaluation on four benchmark datasets}
\label{sec:baseline}
We then compare our proposal with competitive methods on four datasets when the poison ratio is greater than or equal to 40\%, and the perturbation size is 25/255. Table \ref{Table:baseline} shows the classification accuracy when the model is trained on a mixture of perturbed and clean data when the poison ratio varies from 40\% to 100\%. The 11th row (Ours-Full) in Table \ref{Table:baseline} shows the attack performance of our proposal in the full knowledge setting, which assumes the clean data used by the data exploiter is known for the data owner. Compared to other methods, our approach results in at least a 20\%, 15\%, and 10\% greater decrease in the classification accuracy when the poison ratio is 80\%, 60\%, and 40\%, respectively. This demonstrates the superiority of our proposal when the model is trained on a mixture of clean and poisoned data. The 12th row (Ours-Oracle) in Table \ref{Table:baseline} shows our proposal in the sampling oracle setting, which assumes the clean data used by the data exploiter is unknown to the data owner. The performance achieved by our proposal in this setting is close to those achieved in the full knowledge setting. This indicates that our proposal can work well in a more realistic setting. When the poison ratio is 100\% (the last column in each dataset of Table \ref{Table:baseline}), our proposal achieves competitive results compared to previous methods. 

\begin{table*}[t]
\caption{Results of the partial (poison ratio=80-40\%) availability attack in the sampling oracle setting with various surrogate models and target models. The attack performance was evaluated with classification accuracy (\%) on CIFAR-10 datasets. Lower means better attack performance.}
\centering
\scalebox{0.85}{\begin{tabular}{l|lll|lll|lll|lll}
\hline
Target model    & \multicolumn{3}{c|}{Res-18} & \multicolumn{3}{c|}{Res-50} & \multicolumn{3}{c|}{VGG16} & \multicolumn{3}{c}{DenseNet121} \\ \hline
clean accuracy  & \multicolumn{3}{c|}{94.1}   & \multicolumn{3}{c|}{94.2}   & \multicolumn{3}{c|}{90.8}  & \multicolumn{3}{c}{95.2}        \\ \hline
surrogate/ratio & 40      & 60      & 80      & 40      & 60      & 80      & 40      & 60      & 80     & 40        & 60       & 80       \\ \hline
ConvNet         & 72.0    & 64.1    & 60.2    & 76.9    & 70.3    & 62.6    & 70.9    & 61.0    & 54.8   & 68.7      & 62.4     & 55.5     \\ \cline{1-1}
Res-18          & 74.4    & 68.5    & 62.0    & 72.5    & 74.2    & 67.8    & 69.4    & 64.2    & 62.3   & 78.3      & 76.3     & 68.6     \\ \cline{1-1}
VGG11           & 73.9    & 70.9    & 65.2    & 74.6    & 71.8    & 73.6    & 68.8    & 57.6    & 59.5   & 80.1      & 78.9     & 70.2     \\ \hline
\end{tabular}}
\label{table: diff_arch}
\end{table*}

\subsection{Evaluation across different model architectures}
\label{sec:diff_arch}
In practice, the data owner often does not know the exact structure of the target model used by the data exploiter, leading to situations where the surrogate model's architecture differs from that of the target model. We evaluate how our proposal performs in this situation in Table \ref{table: diff_arch} on CIFAR-10. In this experiment, we used three different surrogate models VGG16 \cite{vgg}, ResNet-18 \cite{resnet}, and a three-layer ConvNet to generate perturbations. We train the classification model by four different models VGG16, ResNet-18, ResNet-50 \cite{resnet}, DenseNet-121 \cite{densenet}. Then, we train models on a mixture of clean and poisoned data and evaluate the classification accuracy of these models. In this experiment, we followed the sampling oracle setting. Overall, in Table \ref{table: diff_arch}, we can observe that with all three surrogate models, there is a performance degradation of the different target models. This demonstrates our proposal can work well in a realistic situation where the structure of the surrogate model does not match that of the target model.

We can find that when the surrogate model is ConvNet (the 4th row in Table \ref{table: diff_arch}), the proposed method achieves the highest performance degradation. Especially when the target model is DenseNet-121, the use of ConvNet exhibits a clear advantage over the other two surrogate models. We speculate that this is because, with fewer model parameters, it is easier to reduce the parameter distance between models through the optimization of perturbations.

\begin{table*}[t]
\caption{Results of the full (poison ratio=100\%) and partial (poison ratio=80-40\%) availability attack in the full knowledge setting (Ours-Full) and sampling oracle setting (Ours-Oracle) on CIFAR-10 against countermeasures. The attack performance was evaluated with classification accuracy (\%) for static and partial adaptive countermeasures. Lower means better attack performance. We evaluated the detection rate for full adaptive countermeasures.}
\centering
\scalebox{0.8}{\begin{tabular}{c|c|cccc|cccc}
\hline
                        &                             & \multicolumn{4}{c|}{Ours-Full} & \multicolumn{4}{c}{Ours-Oracle} \\ \hline
                        & Countermeasure/poison ratio & 40\%  & 60\%  & 80\%  & 100\%  & 40\%   & 60\%  & 80\%  & 100\%  \\ \hline
\multirow{5}{*}{Static} & w/o                         & 71.5  & 63.8  & 57.8  & 14.8   & 72.0   & 64.1  & 60.2  & 19.4   \\
                        & Cutout                      & 71.3  & 61.4  & 53.9  & 15.1   & 70.9   & 63.2  & 58.7  & 20.1   \\
                        & Cutmix                      & 73.6  & 66.5  & 58.9  & 22.3   & 74.2   & 66.8  & 63.5  & 24.3   \\
                        & Mixup                       & 73.9  & 68.7  & 60.4  & 25.7   & 75.0   & 66.3  & 67.1  & 25.1   \\
                        & Adv training                & 73.4  & 69.6  & 65.4  & 43.9   & 74.1   & 69.4  & 63.4  & 44.9   \\ \hline
Partial Adaptive        & Orthogonal Projection       & 58.7  & 54.2  & 48.9  & 19.2   & 57.3   & 55.3  & 47.2  & 20.6   \\ \hline
 Full Adaptive& Poisoned source detection& \multicolumn{8}{c}{Detection rate: 100\%}\\ \hline

\end{tabular}}
\label{defense_table}
\end{table*}

\subsection{Evaluation against countermeasure}
The data exploiter may deploy countermeasures to nullify availability attacks. We categorize existing countermeasures for availability attacks into three types: (1) \textbf{Static Countermeasure}: data exploiters are unaware of the availability attack, but they use strong data augmentation techniques such as AugMix \cite{augmix}, Cutmix \cite{yun2019cutmix}, Mixup \cite{mixup} or adversarial training when training the model. These methods have been proven to alleviate the effects of availability attacks \cite{at,shortcut_squ}. The third-seventh row in Table \ref{defense_table} shows that our proposal is still effective against static defense.

(2) \textbf{Partial Adaptive Countermeasure}: data exploiters realize that the data might be protected by an availability attack, but they are unaware of the specific method employed by the data owner. Consequently, they utilize general countermeasures specifically designed to nullify availability attacks. The most direct countermeasure is to add a small amount of clean data to the training set. 
Because the current belief is that availability attacks can be nullified with just 20\% clean data, as shown in Table \ref{defense_table}. Our method renders this strategy ineffective, as shown in Table \ref{Table:baseline}. Additionally, \cite{whatcanwelearn} proposed a method, termed Orthogonal Projection, aimed at training the model without relying on the most predictive features to defend against availability attacks. The eighth row in Table \ref{defense_table} shows that our proposal is effective against orthogonal projection.

(3) \textbf{Full Adaptive Countermeasure}: data exploiters realize that the data owner has implemented PMA and, therefore, understands that several of the multiple data sources they collected are poisoned. They can design adaptive countermeasures targeting the characteristics of PMA. A simple approach is to detect the poisoned data source. Since data exploiters collect data from both poisoned and clean data sources, they can independently train one model for each data source, resulting in $n$ models if there are $n$ data sources. When the model is trained solely on the poisoned data source, it is equivalent to training on 100\% poisoned data, resulting in very low test accuracy. Conversely, when the model is trained on a clean data source, it achieves much better performance. By leveraging this difference, data exploiters can detect the poisoned data source. They can then remove the data from the poisoned source. Such a strategy can effectively identify the poisoned data sources, as shown in the ninth row in Table \ref{defense_table}, making it an effective way to detect poisoned data sources. However, data exploiters need to spend extra training resources for defense, increasing the consumption of using private data. Moreover, even if poisoned data sources are identified, data exploiters will deprecate these poisoned data sources. For this reason, even if poisoned data are detected with high probability, unwanted use of our data is at least hampered. Detailed settings for applying detection are shown in the supplementary.

\section{Conclusion}
In this paper, we propose a partial availability attack, termed Parameter Matching Attack (PMA). By designing a destination model with low test accuracy, the proposed algorithm aims to generate perturbation so that when a model is trained on a mixture of clean data and poisoned data, the resulting model approaches the destination model. In the evaluation, the proposed method shows superior performance on four benchmarks when the poison ratio is 80\%, 60\%, and 40\%, respectively.

\bibliographystyle{unsrt}  
\bibliography{references}

\end{document}